\documentclass[runningheads]{llncs}

\usepackage[T1]{fontenc}
\usepackage{graphicx}
\usepackage{todonotes}
\usepackage{amsmath}
\usepackage{amssymb}
\usepackage{fancyref}
\usepackage{subfig}
\usepackage{booktabs}
\usepackage{multirow}
\usepackage{pgfplots}
\usepackage{wrapfig}
\usepackage{hyperref}

\usepgfplotslibrary{statistics}
\newcommand{\ra}[1]{\renewcommand{\arraystretch}{#1}}

\newcommand\asteriskfill{\leavevmode\xleaders\hbox{$\ast\ $}\hfill\kern0pt}
\newcommand{\comment}[1]{}

\newcommand*\ttvar[1]{\texttt{\expandafter\dottvar\detokenize{#1}\relax}}
\newcommand*\dottvar[1]{\ifx\relax#1\else
  \expandafter\ifx\string_#1\string_\allowbreak\else#1\fi
  \expandafter\dottvar\fi}
  
\pgfplotsset{compat=1.17}

\begin{document}

\title{Bridging the Gap between \\ Deep Learning and Hypothesis-Driven Analysis \\ via Permutation Testing}

\titlerunning{Deep Learning and Hypothesis-Driven Analysis via Permutation Testing}

\author{Magdalini Paschali\inst{1} \and Qingyu Zhao\inst{1}
\and Ehsan Adeli \inst{1} \and
Kilian M. Pohl\inst{1,2}}

\authorrunning{Paschali et al.}

%index{Paschali, Magdalini}
%index{Zhao, Qingyu}
%index{Adeli, Ehsan}
%index{Pohl, Kilian M.}

%tocauthor{Magdalini Paschali, Qingyu Zhao, Ehsan Adeli, Kilian M. Pohl}

\institute{Department of Psychiatry and Behavioral Sciences, Stanford University School of Medicine, Stanford, CA, USA. \and
Center for Health Sciences, SRI International, Menlo Park, CA, USA.
\\\href{mailto:kpohl@stanford.edu}{kpohl@stanford.edu}
}
\maketitle

\begin{abstract}

A fundamental approach in neuroscience research is to test hypotheses based on neuropsychological and behavioral measures, i.e., whether certain factors (e.g., related to life events) are associated with an outcome (e.g., depression). In recent years, deep learning has become a potential alternative approach for conducting such analyses by predicting an outcome from a collection of factors and identifying the most ``informative'' ones driving the prediction. However, this approach has had limited impact as its findings are not linked to statistical significance of factors supporting hypotheses. In this article, we proposed a flexible and scalable approach based on the concept of permutation testing that integrates hypothesis testing into the data-driven deep learning analysis.
We apply our approach to the yearly self-reported assessments of 621 adolescent participants of the National Consortium of Alcohol and Neurodevelopment in Adolescence (NCANDA) to predict negative valence, a symptom of major depressive disorder according to the NIMH Research Domain Criteria (RDoC). Our method successfully identifies categories of risk factors that further explain the symptom.

\keywords{Permutation Testing, Risk Factor Identification, Classification, Behavioral Data, Outcome Prediction, Disease Prediction}
\end{abstract}

\section{Introduction}

Neuropsychological studies often collect a wide range of measurements by asking participants to fill out self-reports and undergo cognitive assessments~\cite{Gurd_2010} in order to gain insights into the intervention and prevention of mental diseases. To support hypotheses motivating study creation, they then select a few measurements and test the statistical significance of their associations with the disease~\cite{Blakesley2009}. Alternatively, deep neural networks (DNNs) can be trained on all collected measurements to predict disease outcomes, and the decision process can be interpreted by identifying critical factors contributing to the prediction~\cite{kang2019}. The identification of such factors is generally based on `importance' scores, i.e., relative measurements (of arbitrary units)~\cite{selvaraju2017grad}. Failing to provide an absolute metric of significance, findings from deep models generally fail to support hypotheses and hence have limited impact on advancing neuropsychology. To bridge the gap between hypothesis-driven analysis and data-driven deep learning, we propose a procedure for testing whether a category (domain) of factors significantly drives the prediction of a DNN. We do so by constructing the distribution of prediction accuracy under the null hypothesis that the tested category does not contain useful information for prediction. 

This null distribution is derived based on a permutation procedure, which has been used to analyze different characteristics of machine learning models. Golland et al. ~\cite{golland2005permutation} relied on permutation analysis to test whether the observed accuracy of a classifier could be achieved by chance. Other methods~\cite{ojala2010permutation} computed permutation-based p-values to quantify whether a classifier exploits the dependency between input features. Permutation testing has also been used for selecting important attributes over a single or a set of prediction models~\cite{fisher2019all} in decision trees~\cite{frank1998using}, random forests~\cite{breiman2001random,altmann2010permutation,ishwaran2019standard}, and DNNs~\cite{mi2021permutation}. Specifically, Mi et al.~\cite{mi2021permutation} proposed a permutation-based feature importance test that, based on normal distributions, identified predictors for survival of kidney-cancer patients.

Compared to these prior approaches, our proposed method has several advantages: 1) our method seamlessly connects data-driven learning approach with traditional hypothesis-driven analysis by quantifying statistical significance of categories of factors; 2) the approach does not require re-training the model as our null hypothesis is linked to a specific trained model; 3) thanks to the non-parametric nature of the permutation test, our method can be adapted to any machine or deep learning model regardless of the accuracy metric used for training.

We applied the proposed procedure to test the significance of categories of neuropsychological and behavioral measurements for predicting the depressive symptom of negative valence of the NIMH Research Domain Criteria (RDoC)~\cite{insel2010research}. To illustrate the generality of our test procedure, we apply it to a cross-sectional and a longitudinal variant of the prediction model, which is trained on annual acquired records of 621 participants (ages 12 to 17 years) provided by the National Consortium on Alcohol and Neurodevelopment in Adolescence (NCANDA). In both scenarios, our permutation procedure identified meaningful categories distinguishing non-symptomatic youth from participants with symptoms of negative valence.
\section{Methodology}

\textbf{Problem Setup} Let $X\in \mathbb{R}^{n\times m}$ be the data matrix recording $m$ measures (capturing demographic, neurospychological, and behavior factors) from $n$ subjects, where $X_i$ denotes the $i$-th row of $X$, associated with subject $i$.
We also assume the $m$ measures can be grouped into $C$ categories so that  $X:=[X^1,...,X^C]$. Each category $X^j\in\mathbb{R}^{n\times m_j}$ consists of $m_j$ measures assessing one specific domain of cognition or behavior (e.g., all measures associated with sleep) for the $n$ subjects ($\sum_{j=1}^C m_j = m$). Furthermore, each subject is linked to a label $y_i$, where $y_i=0$ refers to non-symptomatic (healthy) subjects and $y_i=1$ to a subject of the cohort of interest (e.g., certain disease or negative valence). 

To find the association between factors $X$ and outcome $y$, an emerging approach is to build a deep neural network $f(\cdot)$ so that $y_i'=f(X_i)$. After deriving the prediction $y'=[y_1',...,y_n']^\top$ based on proper cross-validation, an accuracy metric $\psi=\mathcal{M}(y,y')$ is then computed to quantify the predictive power of the model. Examples for $\mathcal{M}(\cdot)$ are the classification accuracy, F1, AUC, mean absolute error, and R2 coefficient. To better understand the complex relationship between brain and behavior relationships associated with the group of interest, the last and yet most important step of the deep learning analysis is to identify which specific categories of factors are significantly associated with the outcome.

\begin{figure}[t]
\centering
  \includegraphics[trim=80 150 100 130, clip,width=\linewidth]{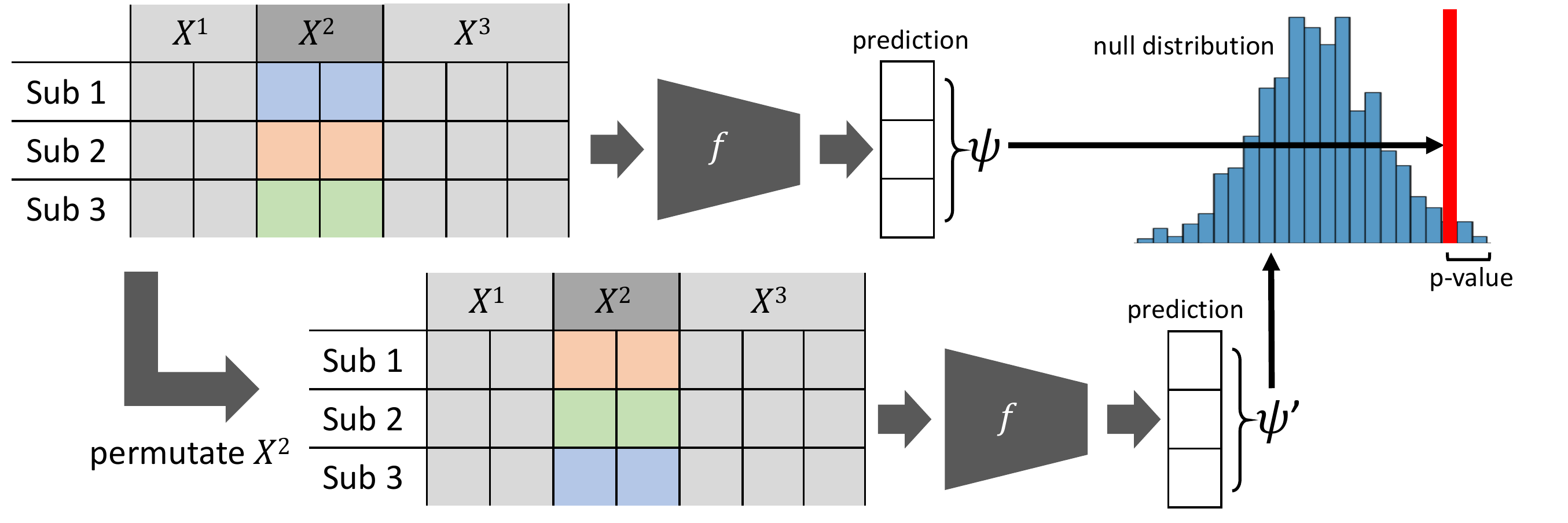}
  \caption{Overview of the proposed permutation method, which we use to compute the significance of a category $X^2$ of factors for driving the prediction of model $f$. We first permute the values of $X^2$ across subjects (rows) and then measure the prediction accuracy based on the permuted input to derive a null distribution of the accuracy score $\hat{\psi}$. The percentile of the true accuracy $\psi$ then defines the category's statistical significance (or p-value).}
  \label{fig:permutation_intro}
\end{figure}

\noindent \textbf{Permutation Test} Instead of using typical model interpretation techniques (such as Gradient-weighted Class Activation Maps (Grad-CAM)~\cite{selvaraju2017grad}) to compute ``importance'' of factors (which is a relative score), we formulate the problem as a hypothesis testing procedure. To test whether a category $X^j$ significantly drives the prediction of the network $f$, our null hypothesis is that 

\textit{$\mathcal{H}_0$: The accuracy of the prediction model $f$ is the same as $\psi$ after permutation of the information in $X^j$.}

Since both the network $f$ and the accuracy metric function $\mathcal{M}$ are possibly highly non-linear functions, we propose to use the permutation test as the test procedure. A permutation test is a non-parametric statistical test that constructs the null distribution of any test statistic ($\psi$ in our case) by calculating all possible values of the test statistic under all possible rearrangements of the observed data points (see also Fig.~\ref{fig:permutation_intro}). Specifically, let $\pi(\cdot)$ denote a random permutation of $n$ objects such that $\pi(X^j)$ denotes the row-permutated matrix for the $j^{th}$ factor category. A new accuracy score is then computed as 
\begin{equation}
\hat{\psi}=\mathcal{M}(f([X^1,...,\pi(X^j),...,X^C]),y).
\end{equation}
In other words, a random permutation across subjects ensures the permutated category no longer carries information stratifying subjects with respect to $y$. As such, repeating the permutation for a large number of trials would result in a distribution of $\hat{\psi}$, i.e., the distribution of the accuracy score under the null hypothesis. Lastly, a p-value can be derived by the percentage of permutations that result in higher accuracy scores than the actual accuracy $\psi$. If the p-value is smaller than a threshold (e.g., 0.05), the null hypothesis is rejected, suggesting that the tested category is significantly associated with the outcome. 
\section{Experimental Setup}
\textbf{Depression Symptom Prediction}
In this work, we predict the depression symptom of negative valence among adolescents from their demographics, self-reports, and neuropsychological and behavioral measures. Negative valence is a symptom describing feelings of sadness, loss, anxiety, fear, and threat. The literature has shown that youth with negative valence tend to develop major depressing disorder (MDD)~\cite{insel2010research} resulting in increased risk for chronic and recurrent depression and suicide attempts.

\noindent\textbf{Dataset}
The NCANDA~\cite{pohl2021ncandarelease} study recruited 831 youths
across five sites in the U.S. (University of California at San Diego (UCSD), SRI International, Duke University Medical Center, University of Pittsburgh (UPMC), and Oregon Health \& Science University (OHSU)) and followed them annually~\cite{brown2015}. As our goal was to analyze adolescent depressive symptoms, we used data from 621 participants who completed at least one assessment before turning 18 years old. The data were part of the public release $\text{\ttvar{NCANDA\_PUBLIC\_6Y\_REDCAP\_V01}}$~\cite{pohl2021ncandarelease}. Among the 621 subjects, 81 reported symptoms of negative valence in at least one of their assessments. These subjects had, on average, 3.20 $\pm$ 1.66 assessments collected every 1.05 $\pm$ 0.15 years. 310 subjects were female and 311 male with an average age of 15.02 $\pm$ 1.69 at the baseline assessment.

We used a total of $m=126$ measurements at each assessment and grouped them into the following categories ($C=8$) according to their content:

\begin{itemize}
    \item \textit{Personality}, which includes traits such as extraversion, agreeableness, acceptance, and emotion regulation.
    \item \textit{Life}, which describes life events, emotional neglect and trauma.
    \item \textit{Sleep}, which includes information regarding a subject's sleep and wake-up time and sleep duration.
    \item \textit{Support}, which describes the involvement of a subject in social clubs and their relationship with their family, friends, and teachers.
    \item \textit{Neuropsych}, which includes emotion recognition, attention, and working memory measures.
    \item \textit{Substance Use}, which corresponds to alcohol and substance use, marijuana dependence, and history.
    \item \textit{Demographics}, which includes age, sex, ethnicity, and pubertal development score.
    \item \textit{BRIEF}, or the Behavior Rating Inventory of Executive Function~\cite{gioia2000brief}, which measures aspects of executive functioning.
\end{itemize}

\noindent \textbf{Model Training}
To showcase our testing approach can be applied to any deep neural network, we trained a cross-sectional and a longitudinal model. Specifically, for the cross-sectional setting, we computed the average value of each feature across all assessments of a subject. Afterward, we classified subjects as having negative valence or being non-symptomatic via a deep learning model consisting of three Fully Connected (FC) layers, ReLU activation, and dropout layers with probability of $0.2$. The model was trained for 55 epochs with a learning rate of 0.001.

For the longitudinal setting, a deep learning model consisting of a Gated Recurrent Unit (GRU)~\cite{gru} layer, and two FC layers identified negative valence in the last assessment of a subject, i.e., in a Sequence-to-One fashion. The model was trained for 30 epochs with learning rate of 0.0001 and Adam Optimizer~\cite{kingma2014adam}.

The loss function used in both settings was the weighted binary cross entropy~\cite{murphybook}. 
To combat the severe class imbalance, we calculated the ratio of non-symptomatic to symptomatic  subjects, $R = \frac{N_{\text{non-symptomatic}}}{N_{\text{symptomatic}}}$ as our loss weight. Additionally, we employed $\ell_1$ regularization to avoid overfitting. We implemented our models and our evaluation in PyTorch 1.10.0~\cite{pytorch}. Our code is publicly available~\footnote{\url{https://github.com/MaggiePas/Permutations_MICCAIPRIME2022}}.

\begin{table}[t]\centering\ra{1.3}
\caption{p-value of each category in predicting negative valence and the null distribution of BACC resulting from category-specific permutations.
Bold marks p-values of significant importance (p$<$0.05). Cross-Sectional refers to models trained on the average across each subject's assessments, and Longitudinal to models trained on all yearly subject assessments.}
\resizebox{0.78\textwidth}{!}{
\begin{tabular}{lcccc} \noalign{\hrule height 0.7pt}
\multicolumn{1}{c}{} & \multicolumn{2}{|c}{\textbf{Cross-Sectional}} & \multicolumn{2}{|c}{\textbf{Longitudinal}} \\
\textbf{Category} & \multicolumn{1}{|c}{\textbf{Null Distribution}} & \textbf{p-value}
& \multicolumn{1}{|c}{\textbf{Null Distribution}} & \textbf{p-value} \\ \hline
Personality & \multicolumn{1}{|c}{65.72 $\pm$ 0.085} &  \textbf{<0.002} & \multicolumn{1}{|c}{58.72 $\pm$ 0.065} & \textbf{<0.002}
 \\
Life & \multicolumn{1}{|c}{69.84 $\pm$ 0.092} & \textbf{0.036} & \multicolumn{1}{|c}{70.22 $\pm$ 0.068} & \textbf{0.030}
\\ 
BRIEF & \multicolumn{1}{|c}{69.35 $\pm$ 0.094} & \textbf{0.044} & \multicolumn{1}{|c}{72.29 $\pm$ 0.079} & 0.074\\
Support & \multicolumn{1}{|c}{69.89 $\pm$ 0.071} & 0.068 & \multicolumn{1}{|c}{71.05 $\pm$ 0.096} & \textbf{0.050}
 \\
Sleep & \multicolumn{1}{|c}{70.48 $\pm$ 0.094} & 0.108 & \multicolumn{1}{|c}{71.95 $\pm$ 0.111} & 0.094
 \\
Neuropsych & \multicolumn{1}{|c}{70.49 $\pm$ 0.083}  & 0.116 & \multicolumn{1}{|c}{74.07 $\pm$ 0.090} & 0.430
\\
Substance Use & \multicolumn{1}{|c}{72.39 $\pm$ 0.098}  & 0.534 & \multicolumn{1}{|c}{74.13 $\pm$ 0.095} & 0.362
 \\
Demographics & \multicolumn{1}{|c}{71.84 $\pm$ 0.093} & 0.366 & \multicolumn{1}{|c}{78.04 $\pm$ 0.112} & 0.992
\\
\noalign{\hrule height 0.7pt}\\
\end{tabular}
}
\label{tab:permutation_results}
\end{table}

\noindent \textbf{Identifying Important Feature Categories}
We trained all models using stratified 5-fold cross-validation with subject-level splits across folds. The accuracy of each model was determined via the F1-Score and the balanced accuracy (BACC)~\cite{balancedaccuracy}, which account for the severe class imbalance. We used the proposed approach to test whether each of the 8 categories significantly contributed to the observed BACC. To do so, we performed 500 permutations within each test fold (a total of 2500 permutations) to generate the null distribution of BACC.
\section{Results and Discussion}
We now present and discuss the results of our models for the cross-sectional and longitudinal training setting. First, we review the significant categories identified by the permutation analysis. Afterward, we test for the specificity of our approach by re-training the models once using only significant categories and once only non-significant categories. Subsequently, we calculate the Shapley values~\cite{suhara2017deepmood} for each category feature and show their overlap with the significant categories identified by our permutation analysis. Finally, we perform hierarchical hypothesis testing by dividing the significant feature categories into meaningful subcategories and repeating the permutation experiment on each of them.

\begin{figure}[t]
\centering
  \includegraphics[width=\linewidth]{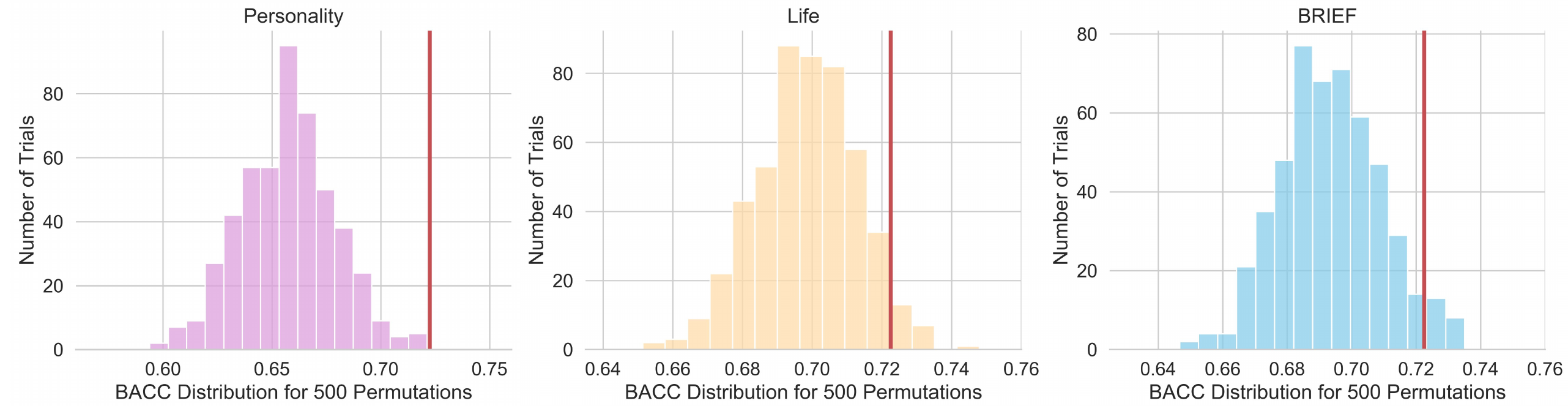}
  \caption{Null distributions of BACC derived by permutation for the three categories that were significant for predicting negative valence when using the cross-sectional model. The red line denotes the observed BACC with the original (un-permutated) data, from which we infer the p-values of the categories as listed in Table 1.}
  \label{fig:perm_dist}
\end{figure}

\noindent \textbf{Cross-sectional Model} Based on the 5-fold cross-validation, the cross-sectional model resulted in a BACC of 72.25\%. 
Table~\ref{tab:permutation_results}(left) summarizes the null distribution of BACC for each category resulting from 500 permutations and the corresponding p-value.  Fig.~\ref{fig:perm_dist} plots the distributions for the significant categories. In all three cases, the mean BACC of the null distribution is clearly lower than the true BACC of 72.25\% (shown in red). 

Specifically, \textit{Personality} is the most significant category for predicting negative valence as permuting personality variables caused the highest drop of 6.5\% (comparing true model BACC with the mean of null distribution). This result was in line with the depression literature, frequently reporting on the connection between  depression and personality traits, which could impact mood through altered reactivity to emotional cues and result in depressive symptoms~\cite{rottenberg2004socioemotional},~\cite{klinger2018predicting},~\cite{watson2015extraversion}.
% Personality traits could also impact mood through altered reactivity to emotional cues resulting in depressive symptoms~\cite{rottenberg2004socioemotional}.
\begin{table}[t]\centering\ra{1.5}
\caption{BACC and F1-scores of models trained with all feature categories, trained only with the categories identified as significant by our approach and only with the non-significant categories. Bold marks the highest BACC and F1-Score.}
\resizebox{0.6\textwidth}{!}{
\begin{tabular}{@{}lcccc@{}} \noalign{\hrule height 0.7pt}
\multicolumn{1}{c}{} & \multicolumn{2}{|c}{\textbf{Cross-Sectional}} & \multicolumn{2}{|c}{\textbf{Longitudinal}} \\	
\textbf{Categories} & \multicolumn{1}{|c}{\textbf{BACC}} & \textbf{F1-Score}	
& \multicolumn{1}{|c}{\textbf{BACC}} & \textbf{F1-Score} \\ \hline
All & \multicolumn{1}{|c}{\textbf{0.7225}} &  \textbf{0.6404} & \multicolumn{1}{|c}{0.7456} & \textbf{0.6482}
 \\		
Only Significant & \multicolumn{1}{|c}{0.7146} & 0.6322 & \multicolumn{1}{|c}{\textbf{0.7668}} & 0.6221
\\ 		
Non-Significant & \multicolumn{1}{|c}{0.6200} & 0.5457 & \multicolumn{1}{|c}{0.6683} & 0.5255\\
\noalign{\hrule height 0.7pt}
\end{tabular}
}
\label{tab:feature_selection}
\end{table}

The second category where random permutations caused a significant decrease in model BACC was \textit{Life}.
%  Discussion Life
It has been consistently documented that youth experiencing more adverse life events are more susceptible to developing depression~\cite{de2013relationship}. Moreover, strong relationships between childhood abuse and depression have also been previously reported~\cite{kendler2000childhood}.
In the cross-sectional setting, \textit{BRIEF} was also a significant predictor for negative valence.
%  Discussion BRIEF
Cognitive and behavioral shifts (that are described by \textit{BRIEF}) have been associated with negative valence, while depression is reciprocally linked with executive dysfunction~\cite{gotlib2010cognition}.

\noindent  \textbf{Longitudinal Model} Our longitudinal model resulted in a 74.56\% BACC. The 2\% increase compared to the cross-sectional model could be attributed to the temporal information provided to our RNN and the higher number of individual assessments.
In line with the cross-sectional setting, \textit{Personality} and \textit{Life} were significant in predicting the outcome. Slightly deviating from the cross-sectional results, permutating \textit{BRIEF} variables only resulted in trend-level significance ($p=0.074$; cross-sectional setting: $p=0.044$) and \textit{Support} (which was at a trend-level $p=0.068$ with respect to the cross-sectional setting) reached the significance threshold. The literature has shown the crucial role of social support in preventing depression~\cite{lamblin2017social}. Moreover, the presence and quality of friendships in adolescence is a significant factor that influences mental health, while limited social interactions could lead to depressive symptoms~\cite{ueno2005effects}. The general agreement in the importance of categories (top 4 vs. bottom 4) between cross-sectional and longitudinal analyses suggests the robustness of our proposed test procedure.

\noindent \textbf{Specificity Testing of Significant Categories} To explore the specificity of our test procedure, we retrained our models using only the significant categories as the input (\textit{Personality}, \textit{Life} and \textit{BRIEF} in the cross-sectional setting and \textit{Personality}, \textit{Life} and \textit{Support} in the longitudinal setting) and also using only the non-significant categories. Table~\ref{tab:feature_selection} shows that results from using only significant categories roughly aligned with the ones using all categories. This indicates that the significant categories identified by our method contained the majority of the information for prediction.

Notably, in the longitudinal setting, the model trained only on the significant categories led to a 2.2\% improvement in BACC, potentially suggesting that removing irrelevant features reduced the chance of overfitting in the RNN. 
In both settings, training the model on the non-significant categories led to a drastic decrease in BACC by 10\% in the cross-sectional model and 8\% in the longitudinal one. These results further highlight that the variables of the non-significant categories did not substantially aid the prediction process. 

\begin{figure}[t]
\centering
  \includegraphics[width=0.95\linewidth]{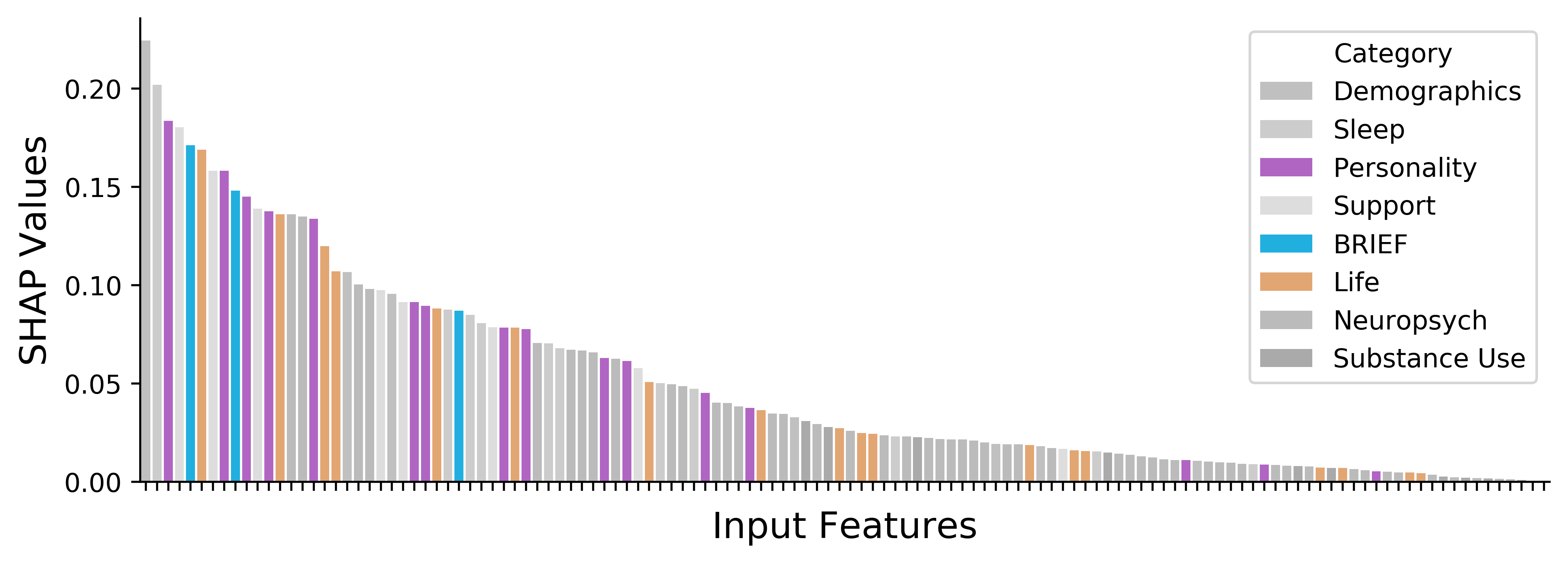}
  \caption{Feature Importance Scores for all our model features computed by the Deep SHapley Additive exPlanation (SHAP)~\cite{lundberg2017unified} derived for our cross-sectional model. The significant categories identified by our approach (Table 1) have also been assigned high SHAP values.}
  \label{fig:shap_values}
\end{figure}

\noindent \textbf{Comparison with SHAP Values} To relate the significance of categories to the importance of features, we generated the Deep SHapley Additive exPlanations (SHAP)~\cite{lundberg2017unified}, which assign an importance value (Shapley values~\cite{shapley1953value}) to each feature for a particular prediction. Shapley values calculate the average marginal contribution of each feature towards the model prediction. Fig.~\ref{fig:shap_values} plots the ranked SHAP values for all 126 features of our model in the cross-sectional setting. Notably, the variables of the categories identified as significant by our approach (i.e., \textit{Personality}, \textit{Life} and \textit{BRIEF}) have also been assigned high SHAP scores.

\noindent  \textbf{Hierarchical Hypothesis Testing} We further show our test procedure's versatility by embedding it into a nested analysis. As \textit{Personality} and \textit{Life} were the two significant categories in both models, we tested the statistical significance of subcategories within each category. Specifically, we split \textit{Personality} into the subcategories of measurements collected by the {\it Ten-Item Personality Inventory} (TIPI)~\cite{gosling2003very}, the {\it Responses to Stress Questionnaire} (RSQ)~\cite{connor2000responses}, and {\it Urgency, Premeditation (lack of), Perseverance (lack of), Sensation Seeking} (UPPS)~\cite{cyders2007integration}. TIPI assesses Extraversion, Agreeableness, Conscientiousness, Emotional Stability, and Openness to Experience. RSQ measures coping and involuntary stress responses, and UPPS measures multiple aspects of impulsive personality. Furthermore, we split \textit{Life} into the measurements associated with Life Events Questionnaire (LEQ)~\cite{masten1994life} and Childhood Trauma Questionnaire (CTQ)~\cite{bernstein1994initial}. LEQ captures life events mostly associated with work and family, and CTQ records physical and emotional abuse and neglect. 

Table~\ref{tab:subcategories} summarizes the outcome when permutating each subcategory, which revealed that \textit{TIPI} is the most significant predictor of negative valence among factors of \textit{Personality} for both settings. These results highlight our approach's ability to scale from more extensive category variables to fine-grained variable subsets. 

\begin{table}[t]\centering\ra{1.5}
\caption{Results of permutation testing on fine-grained feature categories within \textit{Personality} and \textit{Life}. Bold marks p-values of significant importance (p<0.05) for each category sub-scales.}
\resizebox{0.62\textwidth}{!}{
\begin{tabular}{@{}lllcccc@{}} \noalign{\hrule height 0.7pt}
&
\multicolumn{1}{c}{} & \multicolumn{2}{|c}{\textbf{Cross-Sectional}} & \multicolumn{2}{|c}{\textbf{Longitudinal}} \\	
\multicolumn{2}{c}{\textbf{Sub-category}} & \multicolumn{1}{|c}{\textbf{p-value}} & \textbf{Difference}	
& \multicolumn{1}{|c}{\textbf{p-value}} & \textbf{Difference} \\ \hline
\parbox[t]{5mm}{\multirow{3}{*}{\rotatebox[origin=c]{90}{\small{Pesonality}}}} &
TIPI & \multicolumn{1}{|c}{\textbf{0.006}} & -0.033 $\pm$ 0.089 & \multicolumn{1}{|c}{\textbf{<0.002}} & -0.124 $\pm$ 0.056
 \\		
& RSQ & \multicolumn{1}{|c}{0.074} & -0.017 $\pm$ 0.090 & \multicolumn{1}{|c}{0.084} & -0.029 $\pm$ 0.082
\\ 		
& UPPS & \multicolumn{1}{|c}{0.374} & -0.005 $\pm$ 0.103 & \multicolumn{1}{|c}{0.964} & +0.021 $\pm$ 0.089\\ \hline
\parbox[t]{5mm}{\multirow{2}{*}{\rotatebox[origin=c]{90}{\small{Life}}}} & 
LEQ & \multicolumn{1}{|c}{0.098} & -0.017 $\pm$ 0.090 & \multicolumn{1}{|c}{\textbf{0.010}} & -0.049 $\pm$ 0.070\\ 
& CTQ & \multicolumn{1}{|c}{0.452} & -0.000 $\pm$ 0.101 & \multicolumn{1}{|c}{0.864} & +0.011 $\pm$ 0.105\\
\noalign{\hrule height 0.7pt}
\end{tabular}
}
\label{tab:subcategories}
\end{table}

\section{Conclusion}

This paper proposed a flexible and scalable approach that successfully combined hypothesis testing with data-driven deep learning analysis. Based on permutation testing, the approach relates the importance of a category of factors in the data-driven prediction process to p-values used in hypothesis testing. We evaluated our permutation scheme for identifying the major depressive symptom of negative valence from the psycho-social and cognitive factors recorded in 621 NCANCA youths. In line with the literature, personality traits and life events were of significant importance for the predictions performed by the cross-sectional DNN and longitudinal RNN. In summary, our comprehensive analysis can identify potential predictors of negative valence during adolescence that might be important for timely intervention, especially given the increased risk for depression starting with the COVID pandemic~\cite{alzueta2021risk}.

\section*{Acknowledgements} This study was in part supported by the National Consortium on Alcohol and Neurodevelopment in Adolescence (NCANDA) by means of research grants from the National Institute on Alcohol Abuse and Alcoholism (NIAAA) AA021697 (PI: KMP) and AA028840
(PI: QZ). The research was also supported by the Stanford Human-Centered Artificial Intelligence (HAI) Google Cloud Credit (PI: KMP).

\bibliographystyle{ieeetr}
\bibliography{bibfile}

\end{document}